\documentclass[letterpaper, 10 pt, conference]{ieeeconf}  %

\usepackage{color}
\usepackage{bm}
\usepackage{amsfonts}
\usepackage{amsmath}
\usepackage{abraces}
\usepackage{multirow}
\usepackage{pstool}
\usepackage{lipsum}
\usepackage{array}
\usepackage{pstool}

\usepackage[hidelinks]{hyperref}
\usepackage[hyphenbreaks]{breakurl}
\newcommand\es{ES}
\newcommand\ess{ES }

\newcommand\bmd[1]{\bm{#1}}
\newcommand\ev{\bmd{y}}
\usepackage{algorithm}
\usepackage{algpseudocode}
\usepackage{subfigure}
\usepackage{cite}

\makeatletter
\def\BState{\State\hskip-\ALG@thistlm}
\algrenewcommand\ALG@beginalgorithmic{\footnotesize}
\makeatother

\IEEEoverridecommandlockouts                              %

\usepackage{graphics} %
\usepackage{epsfig} %
\title{	\LARGE \bf Automatic LQR Tuning Based on \\ Gaussian Process Global Optimization}

\author{
	Alonso Marco$^{1}$, Philipp Hennig$^{1}$, Jeannette Bohg$^{1}$, Stefan Schaal$^{1,2}$ and Sebastian Trimpe$^{1}$
	\thanks{$^{1}$ Max Planck Institute for Intelligent Systems, T\"ubingen, Germany.}
	\thanks{$^{2}$ Computational Learning and Motor Control Lab at the University of Southern California, Los Angeles, CA, USA.}
	\thanks{E-mails: {\tt\small <firstname>.<lastname>@tuebingen.mpg.de}}
	\thanks{This work was supported by the Max Planck Society, the Max Planck ETH Center for Learning Systems, and a Max Planck Grassroots grant to S. Trimpe and P. Hennig.}
	}

\usepackage{fancyhdr}
\newcommand{\mytitle}{\textbf{Accepted final version.}
To appear in \textit{2016 IEEE International Conference on Robotics and Automation}.\\
\copyright 2016 IEEE. Personal use of this material is permitted. Permission from IEEE must be obtained for all other uses, in any current or future media, including reprinting/republishing this material for advertising or promotional purposes, creating new collective works, for resale or redistribution to servers or lists, or reuse of any copyrighted component of this work in other works.}
\begin{document}
\maketitle

\thispagestyle{fancy}	%
\pagestyle{empty}
\setcounter{section}{0}
\begin{abstract}
This paper proposes an automatic controller tuning framework based on linear optimal control combined with Bayesian optimization. With this framework, an initial set of controller gains is automatically improved according to a pre-defined performance objective evaluated from experimental data. The underlying Bayesian optimization algorithm is Entropy Search, which represents the latent objective as a Gaussian process and constructs an explicit belief over the location of the objective minimum. This is used to maximize the information gain from each experimental evaluation. Thus, this framework shall yield improved controllers with fewer e\-va\-lua\-tions compared to alternative approaches. A seven-degree-of-freedom robot arm balancing an inverted pole is used as the experimental demonstrator. Results of two- and four-dimensional tuning problems highlight the method's potential for automatic controller tuning on robotic platforms.
\end{abstract}
\section{INTRODUCTION}
\label{sec:introduction}
Robotic setups often need fine-tuned controller parameters both at low- and task-levels. Finding an appropriate set of parameters through simplistic protocols, such as manual tuning or grid search, can be highly time-consuming. We seek to automate the process of fine tuning a nominal controller based on performance observed in experiments on the physical plant. We aim for information-efficient approaches, where only few experiments are needed to obtain improved performance.

Designing controllers for balancing systems such as in \cite{trimpe2012balancing} or \cite{mason2014full} are typical examples for such a scenario. Often, one can without much effort obtain a rough linear model of the system dynamics around an equilibrium configuration, for example, from first principles modeling.  Given the linear model, it is then relatively straightforward to compute a stabilizing controller, for instance, using optimal control. When testing this nominal controller on the physical plant, however, one may find the balancing performance unsatisfactory, e.g. due to unmodeled dynamics, parametric uncertainties of the linear model, sensor noise, or imprecise actuation. Thus, fine-tuning the controller gains in experiments on the real system is desirable in order to partly mitigate these effects and obtain improved balancing performance.

We have a tuning scenario in mind, where a limited budget of experimental evaluations is allowed (e.g.\ due to limited experimental time on the plant, or costly experiments).  The automatic tuning shall globally explore a given range of controllers 
and return the best known controller after a fixed number of experiments.
During exploration, we assume that it is acceptable for the controller to fail, for example, because other safety mechanisms are in place \cite{akametalu2014reachability}, or it is uncritical to stop an experiment when reaching safety limits (as is the case in experiment considered herein).
For this scenario, we propose a controller tuning framework extending previous work \cite{trimpe2014self}. 
Therein, a Linear Quadratic Regulator (LQR) is iteratively improved based on control performance observed in experiments. The controller parameters of the LQR design are adjusted using Simultaneous Perturbation Stochastic Approximation (SPSA) \cite{spall2003simultaneous} as optimizer of the experimental cost. It obtains a very rough estimate of the cost function gradient from few cost evaluations, and then updates the parameters in its negative direction. While control performance could be improved in experiments on a balancing platform in \cite{trimpe2014self}, this approach does not exploit the available data as much as could be done. Additionally, rather than exploring the space globally, it only finds local minima.

\begin{figure}[t!]
\centering
\includegraphics[width=\columnwidth]{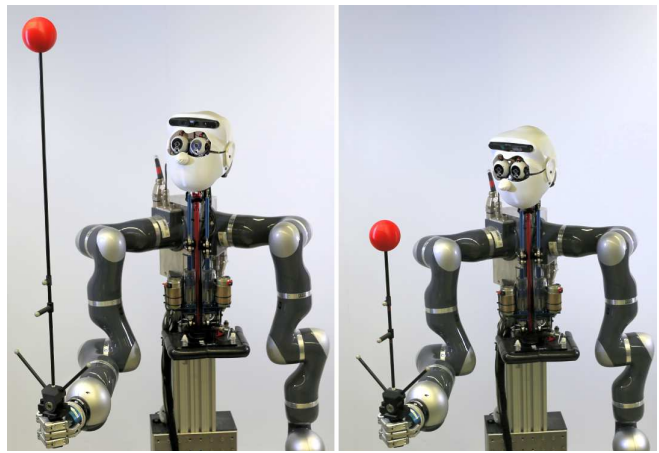}
\caption{The humanoid robot Apollo learns to balance poles of different lengths using the automatic controller tuning framework proposed herein.}
\label{fig:apollo}
\end{figure}

In contrast to \cite{trimpe2014self}, we propose the use of Entropy Search (ES) \cite{hennig2012entropy,villemonteix2009informational}, a recent algorithm for global Bayesian optimization, as the minimizer for the LQR tuning problem. ES employs a Gaussian process (GP) as a non-parametric model capturing the knowledge about the unknown cost function. At every iteration, the algorithm exploits all past data to infer the shape of the cost function. Furthermore, in the spirit of an active learning algorithm, it suggests the next evaluation in order to learn most about the location of the minimum. Thus, we expect ES to be more data-efficient than simple gradient-based approaches as in \cite{trimpe2014self}; that is, to yield better controllers with fewer experiments.

The main contribution of this paper is the development of an automatic controller tuning framework combining ES \cite{hennig2012entropy} with LQR tuning \cite{trimpe2014self}.
 While ES has been applied to numerical optimization problems before, this work is the first to use it for controller tuning on a complex robotic platform.  The effectiveness of the proposed auto-tuning method is demonstrated in experiments of a humanoid robot balancing a pole
 (see Figure \ref{fig:apollo}).  We present successful auto-tuning experiments for parameter spaces of different dimensions (2D and 4D), as well as for initialization with relatively good, but also poor initial controllers.

Preliminary results of this approach are presented in the workshop paper \cite{marco2015irosws}.  
The presentation in this paper is more elaborate and new experimental results are included.
\textit{Related work:} 
Automatic tuning of an LQR is also considered in
\cite{Gr84} and \cite{ClKaMo85}, for example.  In these references, the tuning typically happens by first identifying model parameters from data, and then computing a controller from the updated model. In contrast, we tune the controller gain directly thus bypassing the model identification step.  Albeit we exploit a nominal model in the LQR design, this model is not updated during tuning and merely serves to pre-structure the controller parameters.

Using Gaussian processes (GPs) for automatic controller tuning has recently also been proposed in \cite{schreiter20015,Berkenkamp2016SafeController,calandra2015bayesian,metzen2015ActiveBayesOpt}. In \cite{schreiter20015}, the space of controller parameters is explored by selecting next evaluation points of maximum uncertainty (i.e. maximum variance of the GP). In contrast, ES uses a more sophisticated selection criterion: it selects next evaluation points where the expected information gain is maximal in order to learn most about the global minimum. A particular focus of the method in \cite{schreiter20015} is on safe exploration. For this purpose, an additional GP distinguishing safe and unsafe regions (e.g. corresponding to unstable controllers) is learned.

Safe learning is also the focus in \cite{Berkenkamp2016SafeController}, where the Bayesian optimization algorithm for safe exploration from \cite{sui2015safe} is employed. This work restricts the exploration to controllers that incur a small cost with high probability. The method avoids unsafe controllers and finds the optimum within the safely reachable set of controllers. In contrast, ES explores globally and maximizes information gain in the entire parameter space, regardless of a potentially large costs incurred in an individual experiment.

The authors in \cite{calandra2015bayesian} use Bayesian optimization for learning gait parameters of a walking robot. The gait is achieved using a discrete event controller, and transitions are triggered based on sensor feedback and the learned parameters.
Same as herein, \cite{metzen2015ActiveBayesOpt} also uses Entropy Search for controller tuning, and extends this to contextual policies for different tasks. While \cite{schreiter20015} and \cite{metzen2015ActiveBayesOpt} present simulation studies (balancing an inverted pendulum and robot ball throwing, respectively), \cite{Berkenkamp2016SafeController} and \cite{calandra2015bayesian} demonstrate their algorithms in hardware experiments (quadrocopter and 4-DOF walking robot). To the authors' knowledge, \cite{Berkenkamp2016SafeController} and the work herein are the first to propose and experimentally demonstrate Bayesian optimization for direct tuning of continuous state-feedback controllers on a real robotic platform.

The task of learning a controller from experimental rewards (i.e.\ negative cost) is also considered in the rather large area of reinforcement learning (RL), see \cite{KoBaPe13} for a survey. However, the tools used here (GP-based optimization) differ from the classical methods in RL.
\textit{Outline of the paper:} The LQR tuning problem is described in Sec.~\ref{sec:problem}. The use of \ess for automating the tuning is outlined in Sec.~\ref{sec:es}. The experimental results are presented in Sec.~\ref{sec:results}, and the paper concludes with remarks in Sec.~\ref{sec:conc}.

\section{LQR TUNING PROBLEM}
\label{sec:problem}
In this section, we formulate the LQR tuning problem following the approach proposed in \cite{trimpe2014self}.
\subsection{Control design problem}
\label{ssec:control_des}
We consider a system that follows a discrete-time non-linear dynamic model
\begin{equation}
\bmd{x}_{k+1} = \bmd{f}(\bmd{x}_k,\bmd{u}_k,\bmd{w}_k)
\label{eq:non_lin_sys}
\end{equation}
with system states $\bmd{x}_k\in \mathbb{R}^{n_x}$, control input $\bmd{u}_k\in \mathbb{R}^{n_u}$, and zero-mean process noise $\bmd{w}_k$ at time instant $k$. We assume that (\ref{eq:non_lin_sys}) has an equilibrium at $\bmd{x}_k=\bmd{0}$, $\bmd{u}_k=\bmd{0}$ and $\bmd{w}_k=\bmd{0}$, which we want to keep the system at. We also assume that $\bmd{x}_k$ can be measured and, if not, an appropriate state estimator is used.

For regulation problems such as balancing about an equilibrium, a linear model is often sufficient for control design. Thus, we consider a scenario, where a linear model
\begin{equation}
\tilde{\bmd{x}}_{k+1}=\bmd{A}_{\text{n}}\tilde{\bmd{x}}_k + \bmd{B}_{\text{n}}\bmd{u}_k + \bmd{w}_k
\label{eq:nom_model}
\end{equation}
is given as an approximation of the dynamics (\ref{eq:non_lin_sys}) about the equilibrium at zero. We refer to (\ref{eq:nom_model}) as the \textit{nominal model}, while (\ref{eq:non_lin_sys}) are the true system dynamics, which are unknown.

A common way to measure the performance of a control system is through a quadratic cost function such as 
\begin{equation}
J = \lim_{K\rightarrow\infty}\dfrac{1}{K}\mathbb{E}\left[ \sum_{k=0}^{K-1} \bmd{x}_k^T\bmd{Q}\bmd{x}_k + \bmd{u}_k^T\bmd{R}\bmd{u}_k \right] 
\label{eq:cost_def}
\end{equation}
with positive-definite weighting matrices $\bmd{Q}$ and $\bmd{R}$, and $\mathbb{E}\left[ \cdot \right] $ the expected value. The cost (\ref{eq:cost_def}) captures a trade-off between control performance (keeping $\bmd{x}_k$ small) and control effort (keeping $\bmd{u}_k$ small).%

Ideally, we would like to obtain a state feedback controller for the non-linear plant (\ref{eq:non_lin_sys}) that minimized (\ref{eq:cost_def}). Yet, this non-linear control design problem is intractable in general. Instead, a straightforward approach that yields a locally optimal solution is to compute the optimal controller minimizing (\ref{eq:cost_def}) for the nominal model (\ref{eq:nom_model}). This controller is given by the well-known Linear Quadratic Regulator (LQR) \cite[Sec. 2.4]{anderson2007optimal}
\begin{equation}
\bmd{u}_k = \bmd{F}\bmd{x}_k
\label{eq:static_gain}
\end{equation}
whose static gain matrix $\bmd{F}$ can readily be computed by solving the discrete-time infinite-horizon LQR problem for the nominal model $(\bmd{A}_\text{n},\bmd{B}_\text{n})$ and the weights $(\bmd{Q},\bmd{R})$. For simplicity, we write
\begin{equation}
\bmd{F}=\text{lqr}(\bmd{A}_\text{n},\bmd{B}_\text{n},\bmd{Q},\bmd{R}).
\label{eq:controller_no_par}
\end{equation}

If (\ref{eq:nom_model}) perfectly captured the true system dynamics (\ref{eq:non_lin_sys}), then (\ref{eq:controller_no_par}) would be the optimal controller for the problem at hand. However, in practice, there can be several reasons why the controller (\ref{eq:controller_no_par}) is suboptimal: the true dynamics are non-linear, the nominal linear model (\ref{eq:nom_model}) involves parametric uncertainty, or the state is not perfectly measurable (e.g. noisy or incomplete state measurements). While still adhering to the controller structure (\ref{eq:static_gain}), it is thus beneficial to fine tune the nominal design (the gain $\bmd{F}$) based on experimental data to partly compensate for these effects. This is the goal of the automatic tuning approach, which is detailed next.

\subsection{LQR tuning problem}
\label{ssec:LQRpar}
Following the approach in \cite{trimpe2014self}, we parametrize the controller gains $\bmd{F}$ in (\ref{eq:static_gain}) as 
\begin{equation}
\bmd{F}(\bmd{\theta})=\text{lqr}(\bmd{A}_\text{n},\bmd{B}_\text{n},\bmd{W}_x(\bmd{\theta}),\bmd{W}_u(\bmd{\theta}))
\label{eq:controller}
\end{equation}
where $\bmd{W}_x(\bmd{\theta})$ and $\bmd{W}_u(\bmd{\theta})$ are \textit{design weights} parametrized in $\bmd{\theta}\in \mathbb{R}^D$, which are to be varied in the automatic tuning procedure. For instance, $\bmd{W}_x(\bmd{\theta})$ and $\bmd{W}_u(\bmd{\theta})$ can be diagonal matrices with $\theta_j >0$, $j=1,\ldots,D$, as diagonal entries.

Parametrizing controllers in the LQR weights $\bmd{W}_x$ and $\bmd{W}_u$ as in \eqref{eq:controller}, instead of varying the controller gains $\bmd{F}$ directly, restricts the controller search space. This restriction is often desirable for practical reasons.  First, we assume that the nominal model (albeit not perfect) represents the true dynamics reasonable well around an equilibrium. In this situation, one wants to avoid controllers that destabilize the nominal plant or have poor robustness properties, which is ensured by the LQR design\footnote{According to classical results in control theory \cite{kalman1964linear} and \cite{KoGoSe12}, any stabilizing feedback controller \eqref{eq:static_gain} that yields a return difference greater one (in magnitude) can be obtained for some $\bmd{W}_x$ and $\bmd{W}_u$ as the solution to the LQR problem. The return difference is relevant in the analysis of feedback loops \cite{anderson2007optimal}, and its magnitude exceeding one means favorable robustness properties. Therefore, the LQR parameterization \eqref{eq:controller} only discards controllers that are undesirable because they destabilize the nominal plant, or have poor robustness properties.}.
Second, further parametrizing $\bmd{W}_x$ and $\bmd{W}_u$ in $\bmd{\theta}$ can be helpful to focus on most relevant parameters or to ease the optimization problem. While, for example, a restriction to diagonal weights $\bmd{W}_x$ and $\bmd{W}_u$ is common practice in LQR design (i.e. $n_x + n_u$ parameters), it is not clear how one would reduce the dimensionality of the gain matrix $\bmd{F}$ ($n_x\times n_u$ entries) when tuning this directly. We expect this to be particularly relevant for high-dimensional problems, such as control of a full humanoid robot \cite{mason2014full}.

When varying $\bmd{\theta}$, different controller gains $\bmd{F}(\bmd{\theta})$ are obtained. These will affect the system performance through (\ref{eq:static_gain}), thus resulting in a different cost value from (\ref{eq:cost_def}) in each experiment. To make the parameter dependence of (\ref{eq:cost_def}) explicit, we write
\begin{equation}
J = J(\bmd{\theta}).
\end{equation}

The goal of the automatic LQR tuning is to vary the parameters $\bmd{\theta}$ such as to minimize the cost (\ref{eq:cost_def}).

\textit{Remark:} The weights $(\bmd{Q},\bmd{R})$ in (\ref{eq:cost_def}) are referred to as \textit{performance weights}. Note that, while the \textit{design weights} $\left( \bmd{W}_x(\bmd{\theta}),\bmd{W}_u(\bmd{\theta})\right) $ in (\ref{eq:controller}) change during the tuning procedure, the performance weights remain unchanged.

\subsection{Optimization problem}
\label{ssec:opti}
The above LQR tuning problem is summarized as the optimization problem
\begin{equation}
\arg\min J(\bmd{\theta}) \quad \text{s.t. } \bmd{\theta} \in \mathcal{D}
\label{eq:min_fun}
\end{equation}
where we restrict the search of parameters to a bounded domain $\mathcal{D}\subset \mathbb{R}^D$.
The domain $\mathcal{D}$ typically represents a region around the nominal design, where performance improvements are to be expected or exploration is considered to be safe.

The shape of the cost function in (\ref{eq:min_fun}) is unknown. Neither gradient information is available nor guarantees of convexity can be expected. Furthermore, (\ref{eq:cost_def}) cannot be computed from experimental data in practice as it represents an infinite-horizon problem. As is also done in  \cite{trimpe2014self}, we thus consider the approximate cost
\begin{equation}
\hat{J} = \dfrac{1}{K}\left[ \sum_{k=0}^{K-1} \bmd{x}_k^T\bmd{Q}\bmd{x}_k + \bmd{u}_k^T\bmd{R}\bmd{u}_k \right]
\label{eq:cost_approx}
\end{equation}
with a finite, yet long enough horizon $K$. The cost (\ref{eq:cost_approx}) can be considered a noisy evaluation of (\ref{eq:cost_def}). Such an evaluation is expensive as it involves conducting an experiment, which lasts few minutes in the considered balancing application.

\section{LQR TUNING WITH ENTROPY SEARCH}
\label{sec:es}
In this section, we introduce Entropy Search (ES) \cite{hennig2012entropy} as the optimizer to address problem (\ref{eq:min_fun}). The key characteristics of \ess are explained in Sec.~\ref{ssec:gp_theory} to \ref{ssec:loss}, the resulting framework for automatic LQR tuning is summarized in Sec.~\ref{ssec:aLQR}, and Sec.~\ref{ssec:OtherMethodsThanES} briefly discusses related methods.
Here, we present only the high-level ideas of ES from a practical standpoint. The reader interested in the mathematical details, as well as further explanations, is referred to \cite{hennig2012entropy}.

\subsection{Underlying cost function as a Gaussian process}
\label{ssec:gp_theory}
ES is one of several popular formulations of Bayesian Optimization \cite{kushner1964new,jones1998efficient,srinivas2009gaussian,auer2003using,osborne2009gaussian}, a framework for global optimization in which uncertainty over the objective function $J$ is represented by a probability measure $p(J)$, typically a Gaussian process (GP) \cite{rasmussen2006gaussian}. The shape of the cost function (\ref{eq:cost_def}) is unknown; only noisy evaluations (\ref{eq:cost_approx}) are available. 
A GP is a probability measure over a space of functions. It encodes the knowledge we have about the underlying cost function. Additional information about this cost, gathered through experiments (i.e. noisy evaluations of it), is incorporated by conditioning, which is an analytic operation if the evaluation noise is Gaussian; refer to \cite{rasmussen2006gaussian} for more details.

We model prior knowledge about $J$ as the GP
\begin{equation}
J(\bmd{\theta})\sim \mathcal{GP}\left( \mu(\bmd{\theta}),k(\bmd{\theta},\bmd{\theta}_*)\right) 
\label{eq:gp}
\end{equation}
with mean function $\mu(\bmd{\theta})$ and covariance function $k(\bmd{\theta},\bmd{\theta}_*)$.  Common choices are a zero mean function ($\mu(\bmd{\theta})=\bmd{0}$ for all $\bmd{\theta}$), and the squared exponential (SE) covariance function
\begin{equation}
k_\text{SE}(\bmd{\theta},\bmd{\theta}_*) = \sigma^2\exp\left[ -\dfrac{1}{2}(\bmd{\theta}-\bmd{\theta}_*)^\text{T}\bmd{S}(\bmd{\theta}-\bmd{\theta}_*) \right] 
\label{eq:ker}
\end{equation}
which we also use herein.  The covariance function $k(\bmd{\theta},\bmd{\theta}_*)$ captures the covariance between $J(\bmd{\theta})$ and $J(\bmd{\theta}_*)$.  It can thus be used to encode assumptions about properties of $J$ such as smoothness, characteristic length-scales, and signal variance.  In particular, the SE covariance function \eqref{eq:ker} models very smooth functions with signal variance $\sigma^2$ and length-scales $\bmd{S}=\text{diag}(\lambda_1,\lambda_2,\ldots,\lambda_D)$, $\lambda_j>0$.
We assume that the noisy evaluations (\ref{eq:cost_approx}) of (\ref{eq:cost_def}) can be modeled as
\begin{equation}
\hat{J}=J(\bmd{\theta})+\varepsilon
\label{eq:lik}
\end{equation}
with Gaussian noise $\varepsilon$ of variance $\sigma_\text{n}^2$, yielding the likelihood.
To simplify notation, we write $\bmd{\ev} = \{\hat{J}^i\}_{i=1}^N $ for $N$ evaluations at locations $\bmd{\Theta}=\{ \bmd{\theta}^i\}_{i=1}^N $. Conditioning the GP on the data $\left\lbrace \ev,\bmd{\Theta} \right\rbrace $ then yields another GP with posterior mean $\bar{\mu}(\bmd{\theta})$ and a posterior variance $\bar{k}(\bmd{\theta},\bmd{\theta}_*)$.

Figure \ref{fig:gp} provides an example for a one-dimensional cost function and three successive function evaluations.  
As can be seen, the shape of the mean is adjusted to fit the data points, and the uncertainty (standard deviation) is reduced around the evaluations. In regions where no evaluations have been made, the uncertainty is still large. 
We gather the hyperparameters of the GP in the set $\mathcal{H}=\left\lbrace \lambda_1,\lambda_2,\ldots,\lambda_D,\sigma,\sigma_\text{n}\right\rbrace$. 
An initial choice of $\mathcal{H}$ is improved with every new data point $\hat{J}^i$ %
by maximizing the marginal likelihood, a popular approximation.
In addition, we use automatic relevance determination \cite[Sec. 5.1]{rasmussen2006gaussian} in the covariance function (\ref{eq:ker}), which removes those parameter dimensions with low influence on the cost as more data points become available.

\subsection{Probability measure over the location of the minimum}
\label{ssec:p_min}
\begin{figure}[t!]
\centering
\subfigure[1 evaluation]{
\input{./Pics/GP1D/GP_post1.tex}
\includegraphics[width=1\columnwidth]{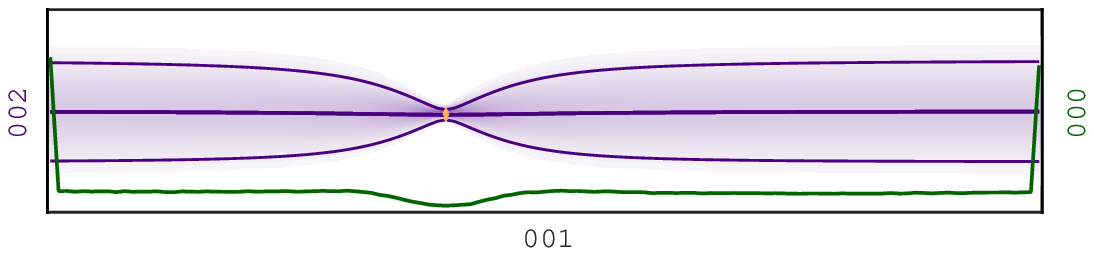}}
\subfigure[2 evaluations]{
\input{./Pics/GP1D/GP_post2.tex}
\includegraphics[width=1\columnwidth]{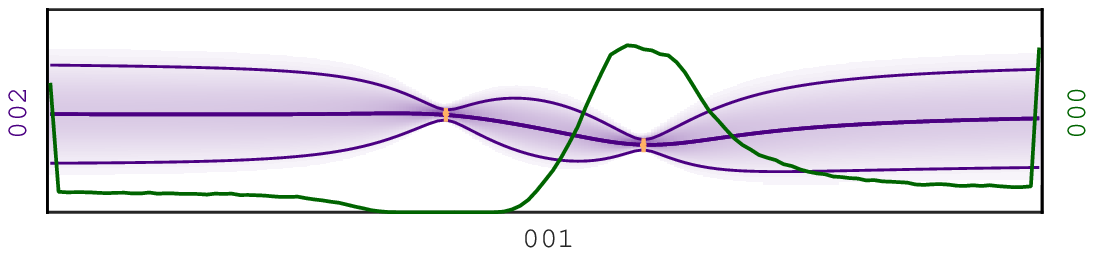}}
\subfigure[3 evaluations]{
\input{./Pics/GP1D/GP_post3.tex}
\includegraphics[width=1\columnwidth]{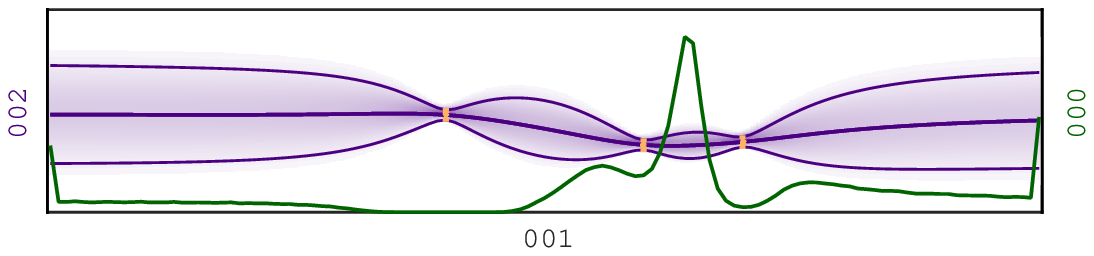}}
\caption{Evolution of an example Gaussian process 
for three successive function evaluations (orange dots), 
reproduced with slight alterations from \cite{hennig2012entropy}.
The posterior mean $\bar{\mu}(\bmd{\theta})$ is shown in thick violet, two standard deviations $2\bar{\sigma}(\bmd{\theta})$ in thin violet, and the probability density as a gradient of color that decreases away from the mean. Two standard deviations of the likelihood noise 2$\sigma_\text{n}$ are represented as orange vertical bars at each evaluation. 
Approximated probability distribution over the location of the minimum $p_\text{min}(\bmd{\theta})$ in green. This plot uses arbitrary scales for each object.}
\label{fig:gp}
\end{figure}

A key idea of ES 
is to explicitly represent the probability $p_\text{min}(\bmd{\theta})$ 
for the minimum location over the domain $\mathcal{D}$:
\begin{equation}
p_\text{min}(\bmd{\theta})\equiv p( \bmd{\theta} = \arg \min J(\bmd{\theta}) ), \quad \bmd{\theta} \in \mathcal{D} .
\label{eq:p_min_rough}
\end{equation}
The probability $p_\text{min}(\bmd{\theta})$ is induced by the GP for $J$:  given a distribution of cost functions $J$ as described by the GP, one can in principle compute the probability for any $\bmd{\theta}$ of being the minimum of $J$. For the example GPs in Fig.~\ref{fig:gp}, $p_\text{min}(\bmd{\theta})$ is shown in green.

To obtain a tractable algorithm, ES approximates $p_\text{min}(\bmd{\theta})$ with finitely many points on a non-uniform grid that puts higher resolution in regions of greater influence.  

\subsection{Information-efficient evaluation decision}
\label{ssec:loss}
The key feature of ES is the suggestion of new locations $\bmd{\theta}$, where \eqref{eq:cost_approx} should be evaluated to learn most about the location of the minimum.  This is achieved by selecting the next evaluation point that maximizes the relative entropy 
\begin{equation}
H = \int_{\mathcal{D}}{p_\text{min}({\bmd{\theta}})\log \dfrac{p_\text{min}({\bmd{\theta}})}{b({\bmd{\theta}})}}\text{d}\bmd{\theta}
\label{eq:entropy}
\end{equation}
between $p_\text{min}(\bmd{\theta})$ and the uniform distribution $b(\bmd{\theta})$ over the bounded domain $\mathcal{D}$.  The rationale for this is that the uniform distribution essentially has no information about the location of the minimum, while a very ``peaked'' distribution would be desirable to obtain distinct potential minima.  This can be achieved by maximization of the relative entropy \eqref{eq:entropy}.  
For this, ES selects next evaluations where the first order expansion $\Delta H(\bmd{\theta})$ of the expected change in (\ref{eq:entropy}) is maximal. In this way, the algorithm efficiently explores the domain of the optimization problem in terms of information gain (cf. \cite[Sec. 2.5]{hennig2012entropy}). Conceptually, the choice of the locations $\bmd{\Theta}$ is made such that ``we evaluate where we expect to learn most about the minimum, rather than where we think the minimum is'' \cite[Sec. 1.1]{hennig2012entropy}.

In addition to suggesting the next evaluation, ES also returns its current \textit{best guess} of the minimum location; that is, the maximum of its approximation to $p_\text{min}(\bmd{\theta})$.

\subsection{Automatic LQR tuning}
\label{ssec:aLQR}
The proposed method for automatic LQR tuning is obtained by combining the LQR tuning framework from Section \ref{sec:problem} with \es; that is, using ES to solve \eqref{eq:min_fun}. At every iteration, ES suggests a new controller (through $\bmd{\theta}$ with \eqref{eq:controller}), which is then tested in an experiment to obtain a new cost evaluation (\ref{eq:cost_approx}). Through this iterative procedure, the framework is expected to explore relevant regions of the cost \eqref{eq:cost_def}, infer the shape of the cost function, and eventually yield the global minimum within $\mathcal{D}$. The automatic LQR tuning method is summarized in Algorithm \ref{alg:aLQR}.

The performance weights $(\bmd{Q},\bmd{R})$ encode the desired performance for the system (\ref{eq:non_lin_sys}). Thus, a reasonable initial choice of the parameters $\bmd{\theta}$ is such that the design weights $\left( \bmd{W}_x(\bmd{\theta}),\bmd{W}_u(\bmd{\theta})\right) $ equal $(\bmd{Q},\bmd{R})$. The obtained initial gain $\bmd{F}$ would be optimal if (\ref{eq:nom_model}) were the true dynamics. After $N$ evaluations, 
ES aims to improve this initial choice, by selecting experiments which are expected to provide maximal information about a better parameter setting.

\begin{algorithm}[t!]
\caption{{\small Automatic LQR Tuning. As its inputs, \textsc{EntropySearch} takes the type of covariance function $k$, the likelihood $l$, a fixed number of evaluations $N$, and data points $\{\bmd{\Theta},\bmd{y}\}$. Alternative stopping criteria instead of stopping after $N$ iterations can be used.
}}
\begin{algorithmic}[1]
\State $\text{initialize } \bmd{\theta}^0 \text{; typically } \bmd{W}_x(\bmd{\theta}^0)=\bmd{Q}\text{, }\bmd{W}_u(\bmd{\theta}^0)=\bmd{R}$
\State $\hat{J}^{0}\gets$ \Call{CostEvaluation}{$\bmd{\theta}^{0}$} \Comment{Cost evaluation}
\State $\{\bmd{\Theta},\bmd{y}\} \gets \{\bmd{\theta}^0,\hat{J}^0\}$
\Procedure{EntropySearch}{$k$,$l$,$N$,$\{\bmd{\Theta},\bmd{y}\}$}
	\For{$i=1\text{ to }N$}
		\State $[ \bar{\mu},\bar{k}] \gets \text{GP}(k,l,\{\bmd{\Theta},\bmd{y}\})$ \Comment{GP posterior}
		\State $p_\text{min} \gets \text{approx\_pmin}(\bar{\mu},\bar{k})$ \Comment{Approximate $p_\text{min}$}
		\State $\bmd{\theta}^{i}\gets \arg\max\Delta H$ \Comment{Next location to evaluate at}
		\State $\hat{J}^{i}\gets$ \Call{CostEvaluation}{$\bmd{\theta}^{i}$} \Comment{Cost evaluation}
		\State $\{\bmd{\Theta},\bmd{y}\} \gets \{\bmd{\Theta},\bmd{y}\}\cup\{\bmd{\theta}^{i},\hat{J}^{i}\}$
		\State $\bmd{\theta}^\text{BG}\gets \arg\max p_\text{min}$ \Comment{Update current ``best guess''}
	\EndFor
\State \Return $\bmd{\theta}^\text{BG}$
\EndProcedure
\Statex
\Function{CostEvaluation}{$\bmd{\theta}$}
\State LQR design: $\bar{\bmd{F}}\gets \text{lqr}(\bmd{A}_\text{n},\bmd{B}_\text{n},\bmd{W}_x(\bmd{\theta}),\bmd{W}_u(\bmd{\theta}))$
\State update control law (\ref{eq:static_gain}) with $\bmd{F}=\bar{\bmd{F}}$
\State perform experiment and record $\{\bmd{x}_k\},\{\bmd{u}_k\}$
\State Evaluate cost: $\hat{J}\gets\frac{1}{K}\left[ \sum_{k=0}^{K-1} \bmd{x}_k^T\bmd{Q}\bmd{x}_k + \bmd{u}_k^T\bmd{R}\bmd{u}_k \right]$
\State \Return $\hat{J}$
\EndFunction
\end{algorithmic}
\label{alg:aLQR}
\end{algorithm}

\subsection{Relation to other GP-based optimizers}
\label{ssec:OtherMethodsThanES}

In addition to the novel ES, there exist a number of Bayesian optimization algorithms based on Gaussian process (GP) measures over the optimization objective. Most of these methods do not retain an explicit measure over the location of the optimum. While ES aims at collecting information about the minimum, these methods directly try to collect small function values (a concept known as \emph{minimizing regret}). 
This strategy is encoded in several different heuristic evaluation utilities, including \emph{probability of improvement} (PI) \cite{kushner1964new}, \emph{expected improvement} (EI) \cite{jones1998efficient} and \emph{upper confidence bound} for GP bandits (GP-UCB) \cite{srinivas2009gaussian,auer2003using}. PI is the probability that an evaluation at a specific point lies below the current best guess, and EI is the expected value by which an evaluation might lie below the current best guess. GP-UCB captures the historically popular notion of ``optimism in the face of uncertainty'' and has the analytic appeal of coming with a theoretical worst-case performance guarantee.

The key difference between ES and these other methods is that they directly try to design experiments that yield increasingly low function values. This is the right strategy in settings where the performance of each individual experiment matters, e.g. the gait of a walking robot is improved online, but it is not allowed to fall. However, in a ``prototyping'' setting, where the sole use of experiments is to learn about a final good design, the numerical result of each experiment is less important than its information content.

The proposed automatic controller tuning framework relies on the ``prototyping'' setting, for which each experiment should be as informative as possible about the global minimum of the cost function.

One minor downside of ES is that it has higher computational cost than alternative methods, taking several seconds to decide on the next experiment. However, in our setting, where the physical experiments take significantly longer than this time, this is not a major drawback.

\section{EXPERIMENTAL RESULTS}
\label{sec:results}

In this section, we present auto-tuning experiments for learning to balance a pole as shown in Fig. \ref{fig:apollo}.
A video demonstration that illustrates the second experiment described in Sec.~\ref{ssec:results_2D} is available at \burl{https://am.is.tuebingen.mpg.de/publications/marco_icra_2016}.

\subsection{System description}
We consider a one-dimensional balancing problem: a pole linked to a handle through a rotatory joint with one degree of freedom (DOF) is kept upright by controlling the acceleration of the end-effector of a seven DOF robot arm (Kuka lightweight robot). Figure \ref{fig:apollo} shows the setup for two poles of different length.
The angle of the pole is tracked using an external motion capture system.

The continuous-time dynamics of the balancing problem (similar to \cite{schaal1997learning}) are described by:
\begin{align}
& mr^2\ddot{\psi}(t)-mgr\sin \psi(t)+mr\cos \psi(t)u(t)+\xi\dot{\psi}(t) = 0 \nonumber \\
& \ddot{s}(t) = u(t)
\label{eq:pole_dyn}
\end{align}
where $\psi(t)$ is the pole angle with respect to the gravity axis, $s(t)$ is the deviation of the end-effector from the zero position, and $u(t)$ is the end-effector acceleration.

Two poles with different lengths are used in the experiments.
The center of mass of the short pole lies at $r\simeq 0.33$ m from the axis of the rotatory joint, its mass is $m\simeq 0.27$ kg, the 
friction coefficient is $\xi\simeq 0.012$ Nms, and the gravity constant is $g=9.81$ m/s$^\text{2}$. 
For the long pole, we have $r\simeq 0.64$ m and $m\simeq 0.29$ kg.

A model (\ref{eq:nom_model}) of the system is obtained by linearization of (\ref{eq:pole_dyn}) about the equilibrium $\psi=0$, $s=0$ and discretization with a sampling time of 1 ms. Using the parameters of the short pole, we obtain its nominal model $(\bmd{A}_\text{n},\bmd{B}_\text{n})$.
The non-linear model (\ref{eq:pole_dyn}) assumes that we can command a discretized end-effector acceleration $u_k$ as control input to the system. In reality, this end-effector acceleration is realized through an appropriate tracking controller for the end-effector following a similar control structure as in \cite{righetti2014autonomous}.
The estimated end-effector position $s_k$ and velocity $\dot{s}_k$ are computed at a sampling rate of 1kHz from the robot's joint encoders using forward kinematics. The pole orientation is captured at 200 Hz by the motion capture system.  From this data, we obtain estimates of pole angle $\psi_k$ and angular velocity $\dot{\psi}_k$ through numerical differentiation and low-pass filtering (2nd-order Butterworth, 10 Hz cutoff).  With this scheme, no model is required to obtain estimates of all states (in contrast to the Kalman filter used in \cite{marco2015irosws}), and it can be used irrespective of which balancing pole is used.
The complete state vector of (\ref{eq:nom_model}) is given by $\bmd{x}_k = [ \psi_k, \dot{\psi}_k, s_k, \dot{s}_k]^\text{T}$.

When using a state-feedback controller (\ref{eq:static_gain}) for balancing, biases in the angle measurement 
lead to a steady-state error in the end-effector position (cf. discussion in \cite[p. 67]{trimpe2012balancing} for a similar balancing problem). To compensate for such steady-state offsets, the state feedback controller (\ref{eq:static_gain}) is augmented with an integrator on the end-effector position, which is a standard way to achieve zero steady-state error (see e.g. \cite[Sec. 6.4]{astrom2010feedback}). That is, we implement the control law $u_k = \bmd{F}\bmd{x}_k + F_zz_k$ instead of (\ref{eq:static_gain}), where $z_k$ is the integrator state.

Although $F_z$ can readily be included in the LQR formulation (\ref{eq:controller}) and tuned alongside the other gains (as was done in \cite{marco2015irosws}), we fix $F_z=-0.3$ here for simplicity.  Since the integrator is not a physical state (it is implemented in the controller) and merely affects the long-term behavior, we do not include it in the computation of the cost (\ref{eq:cost_approx}).

\subsection{Automatic LQR tuning: Implementation choices}

We choose the performance weights to be
\begin{equation}
\bmd{Q}=\text{diag}(1,100,10,200),\;\bmd{R}=10
\label{eq:per_weights}
\end{equation}
where $\text{diag}(\cdot)$ denotes the diagonal matrix with the arguments on the diagonal. We desire to have a quiet overall motion in the system. Therefore, we penalize the velocities $\dot{\psi}_k$ and $\dot{s}_k$ more than the position states. 
We conducted two types of tuning experiments, one with two parameters and another one with four. The corresponding design weights are
\begin{itemize}
\item 2D tuning experiments:
\begin{equation}
\bmd{W}_x(\bmd{\theta})=\text{diag}(1,50\theta_1,10,50\theta_2),\; \bmd{W}_u(\bmd{\theta})=10
\label{eq:des_weights_ex_2D}
\end{equation}
where the parameters $\bmd{\theta} = \left[\theta_1,\theta_2  \right]$ can vary in $[0.01, 10]$, and $\bmd{\theta}^\text{0}=\left[2, 4\right] $ is chosen as initial value.

\item 4D tuning experiments:
\begin{equation}
\begin{array}{l}
\bmd{W}_x(\bmd{\theta})=\text{diag}(\theta_1,25\theta_2,10\theta_3,25\theta_4),\\
\bmd{W}_u(\bmd{\theta})=10
\end{array}
\label{eq:des_weights_ex_4D}
\end{equation}
with $\bmd{\theta} = \left[\theta_1,\theta_2,\theta_3,\theta_4  \right]$, $\theta_j \in [0.01, 10]$, and $\bmd{\theta}^\text{0}=\left[1, 4, 1, 8\right] $.
\end{itemize}
In both cases, the initial choice $\bmd{\theta}^\text{0}$ is such that the design weights equal the performance weights. That is, the first controller tested corresponds to the nominal LQR design \eqref{eq:controller_no_par}.

\begin{table}[b!]
\caption{Characterization of the gamma prior over $\mathcal{H}$}
\begin{center}
\begin{tabular}{rm{5ex}m{5ex}||m{5ex}m{5ex}@{}m{0pt}@{}}
& \multicolumn{2}{c||}{2D exploration} & \multicolumn{2}{c}{4D exploration} & \\
\cline{2-5}
& $\mathbb{E}\left[ \cdot \right] $ & $\text{Std}\left[ \cdot \right] $  & $\mathbb{E}\left[ \cdot \right] $ & $\text{Std}\left[ \cdot \right] $  & \\[1ex]
\hline
Lengthscale $\lambda_j$       		 & $2.5$    & $0.11$   & $2.00$  & $0.63$ & \\
Signal variance $\sigma$ 			 & $0.2$    & $0.02$   & $0.75$  & $0.075$ & \\
Likelihood noise $\sigma_\text{n}$  & $0.033$  & $0.0033$ & $0.033$ & $0.010$ & \\
\end{tabular}
\end{center}
\label{tab:gamma_prior}
\end{table}

Balancing experiments were run for 2 minutes, i.e. a discrete time horizon of $K=1.2\cdot 10^5$ steps.  
We start the experiments from roughly the same initial condition. To remove the effect of the transient and slightly varying initial conditions, we omit the first $30\, \text{s}$ from each experiment.

Because the nominal model does not capture the true dynamics, some LQR controllers obtained during the tuning procedure destabilized the system. This means 
that the system exceeded either acceleration bounds or safety constraints on the end-effector position.
In these cases, the experiment was stopped and a fixed heuristic cost $J_\text{u}$ was assigned to the experiment.
Values for $J_\text{u}$ are typically chosen slightly larger than the performance of a stable but poor controller.  We used $J_\text{u}=3.0$ and $J_\text{u}=5.0$ for the 2D and 4D  experiments, respectively.

\begin{figure}[t!]
\centering
\subfigure[ES initialization]{
\includegraphics[width=1\columnwidth]{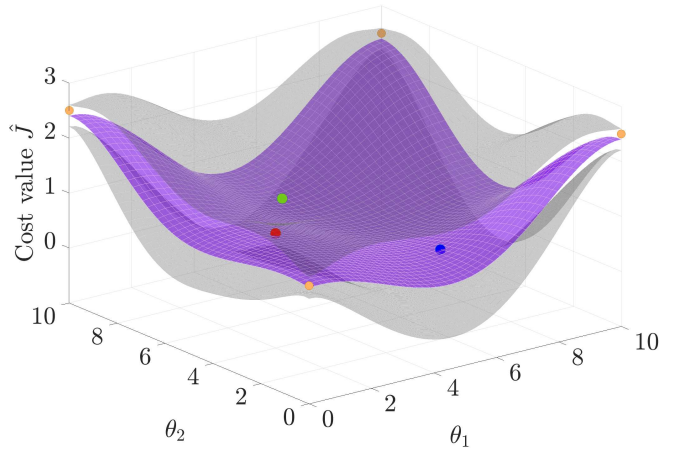}
}
\subfigure[GP posterior after 20 iterations]{
\includegraphics[width=01\columnwidth]{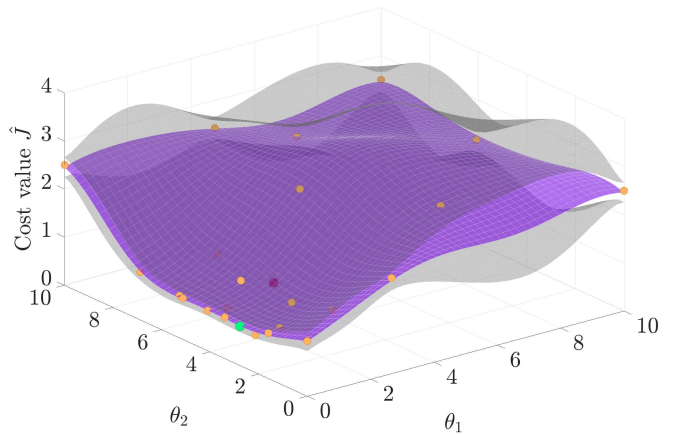}}
\caption{GPs at (a) the start and (b) the end of the first tuning experiment.  The GP mean is represented in violet and $\pm$ two standard deviations in grey. The red dot corresponds to the initial controller, computed at location $\bmd{\theta}^\text{0}=\left[2,4\right] $. The green dot represents the current best guess for the location of the minimum. The blue dot is the location suggested by ES to evaluate next, 
and orange dots represent previous evaluations.
The best guess found after 20 iterations (green dot in (b)) has significantly lower cost than the initial controller (red dot).
}
\label{fig:short_pole_experiments}
\end{figure}

Before running ES, a few experiments were done 
to acquire knowledge about the hyperparameters $\mathcal{H}$. A Gamma prior distribution was assumed over each hyperparameter with expected values and variances shown in Table \ref{tab:gamma_prior}. For the first iteration of ES, we use these expectations as initial set $\mathcal{H}$. After each iteration, $\mathcal{H}$ is updated as the result of maximizing the GP marginal likelihood.
\subsection{Results from 2D experiments}
\label{ssec:results_2D}
For the 2D experiments \eqref{eq:des_weights_ex_2D}, we first use a short pole (Fig.~\ref{fig:apollo}, right) and the best available linear model, showing that the framework is able to improve the initial controller. Secondly, we exchange the pole with one of double length (Fig.~\ref{fig:apollo}, left), but keep the same nominal model. 
We show, for the latter case, that even with a $50\%$ underestimated model, the framework finds a stable controller with good performance. In both cases, we use the design weights (\ref{eq:des_weights_ex_2D}). 
\subsubsection{Using an accurate nominal model}
ES was initialized with five evaluations, i.e. the initial controller $\bmd{\theta}^\text{0}$, and evaluations at the four corners of the domain $[0.01,10]^2$. 
Figure \ref{fig:short_pole_experiments} (a) shows the 2D Gaussian process including the five initial data points.  
The algorithm can also work without these initial evaluations; however, we found that they provide useful prestructuring of the GP and tend to speed up the learning. This way, the algorithm focuses on interesting regions more quickly. 

Executing Algorithm \ref{alg:aLQR} for 20 iterations (i.e. 20 balancing experiments) resulted in the posterior GP shown in Figure \ref{fig:short_pole_experiments} (b).
The ``best guess'' $\bmd{\theta}^\text{BG}=\left[ 0.01,2.80 \right] $ (green dot) is what \ess suggests to be the location of the minimum of the underlying cost (\ref{eq:cost_def}). 
In order to evaluate the result of the automatic LQR tuning, we computed the cost of the resulting controller (best guess after 20 iterations) in five separate balancing experiments. The average and standard deviation of these experiments are shown in Table \ref{tab:results_exp1_2} (left column, bottom), together with the average and standard deviation of the initial controller, computed in the same way before starting the exploration (left column, top). Even though the initial controller was obtained from the best linear model we had, the performance was still improved by $31.9\%$.

\subsubsection{Using a poor nominal model}
\begin{figure}[t!]
\centering
\includegraphics[width=1\columnwidth]{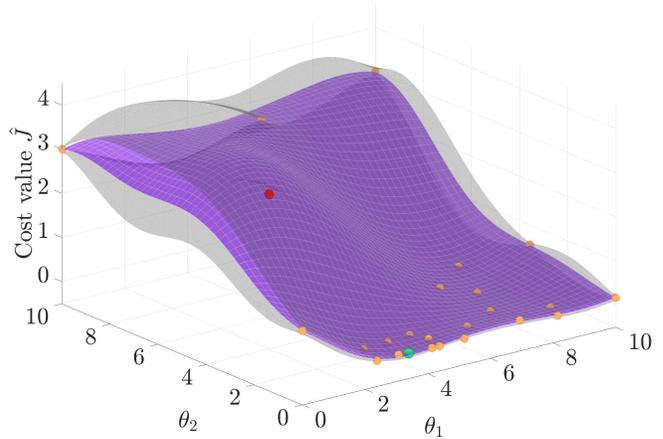}
\caption{Final GP posterior for the second tuning experiment using a wrong nominal model.  The color scheme is the same as in Fig.~\ref{fig:short_pole_experiments}.
}
\label{fig:long_pole_experiments}
\end{figure}

In this experiment, we take the same nominal model as in the previous case, but we use a longer pole in the experimental demonstrator (Fig.~\ref{fig:apollo}, left). The initial controller, computed with $\bmd{\theta}^0$, destabilizes the system, which can be explained by the nominal model significantly misrepresenting the true dynamics. As shown in Figure \ref{fig:long_pole_experiments}, after 20 iterations, ES suggested $\bmd{\theta}^\text{BG}=\left[ 3.25,0.01 \right]$ as the best controller. The results of evaluating this controller five times, in comparison to the initial controller, are shown in Table \ref{tab:results_exp1_2} (middle column). %

\begin{table}[b!]
\caption{Cost values $\hat{J}$ for three tuning experiments} %
\begin{center}
\begin{tabular}{rm{4ex}m{4ex}||m{4ex}m{4ex}||m{4ex}m{3ex}}
& \multicolumn{2}{c||}{2D experiments} & \multicolumn{2}{c||}{2D experiments} & \multicolumn{2}{c}{4D experiments} \\
& \multicolumn{2}{c||}{Good model} & \multicolumn{2}{c||}{Poor model} & \multicolumn{2}{c}{Poor model} \\
\cline{2-7}
& mean & std & mean & std & mean & std \\
\hline
$\bmd{\theta}^\text{0}$   & $1.12$  & $0.11$  & $J_\text{u}$ & - & $J_\text{u}$ & - \\
$\bmd{\theta}^\text{BG}$  & $0.76$  & $0.058$  & $0.059$ & $0.012$ & $0.040$  & $0.0031$ \\
\end{tabular}
\end{center}
\label{tab:results_exp1_2}
\end{table}

\subsection{Results from 4D experiment}
The 4D tuning experiment, realized with the long pole, uses the same nominal model as in the previous experiments (i.e., a poor linear model for the real plant), and the design weights (\ref{eq:des_weights_ex_4D}). We show that the framework is able to improve the controller found during the 2D experiments with the long pole, but in a higher dimensional space.

\begin{figure}[t!]
\centering
\input{./Pics/results_4D/cost.tex}
\includegraphics[width=1\columnwidth]{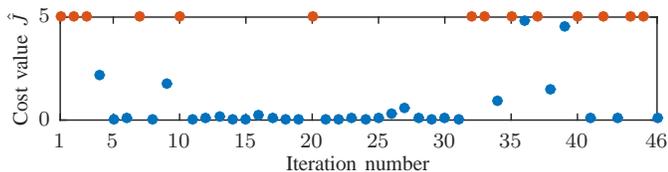}
\caption{Cost values obtained at each experiment distinguishing stable controllers (blue dots), and unstable controllers (red dots). %
}
\label{fig:cost_entropy_bg}
\end{figure}

The first controller $\bmd{\theta}^0$ destabilizes the system. After 46 iterations, ES suggests $\bmd{\theta}^\text{BG}=\left[ 4.21,7.47,0.43,0.01 \right]$, which in comparison with the 2D experiments with the long pole, performs about $31.7\%$ better (see Table \ref{tab:results_exp1_2}).  We actually ran this experiment until iteration 50, however, the algorithm did not lead to further improvements.
Figure \ref{fig:cost_entropy_bg} shows the cost function evaluations over the course of the tuning experiment.
The fact that unstable controllers are obtained throughout the experiment 
reflects how the global search tends to cover all areas. 
Before starting the 2D experiments, we spent some effort selecting the method's parameters, such as hyperparameters and parameter ranges.
In contrast, we started the 4D experiments without such prior tuning.   In particular, we kept the same performance weights, chose similar design weights, and started with the same values for the hyperparameters $\mathcal{H}$ and penalty $J_\text{u}$.
However, we had to restart the search twice in order to slightly adjust $\mathcal{H}$, and $J_\text{u}$.

In general, any method reasoning about functions on continuous domains from a finite number of data points relies on prior assumptions (see \cite[Sec.~1.1]{hennig2012entropy} for a discussion).  
We were quite pleased with the outcome of the tuning experiments and, in particular, that not much had to be changed moving from the 2D to 4D experiment.
Nonetheless, 
developing general rules for choosing the parameters of GP-based optimizers like ES (maybe specific for certain problems) seems important for future developments.

\section{CONCLUDING REMARKS}
\label{sec:conc}
In this paper, we introduce Bayesian optimization for automatic controller tuning.  We develop, and successfully demonstrate in experiments on a robotic platform, a framework based on LQR tuning \cite{trimpe2014self} and Entropy Search (\es) \cite{hennig2012entropy}.  This work is the first to apply ES in experiments for automatic controller tuning.

The auto-tuning algorithm was demonstrated in a 2D and a 4D experiment, both when the method was initialized with an unstable and with a stable controller.  While the 2D experiment could presumably also be handled by grid search or manual tuning, and thus mostly served as a proof of concept, the 4D tuning problem can already be considered difficult for a human.  A key question for the development of truly automatic tuning methods is the amount of ``prior engineering'' that has to be spent to get the method to work.  In particular, the 4D experiments were promising since not a lot of tuning, and only few restarts were necessary.  However, questions pertaining to the prior choice or automatic adjustment of the method's parameters are relevant for future work.
Since the ES algorithm reasons about where to evaluate next in order to maximize the information gain of an experiment, we expect it to make better use of the available data and yield improved controllers more quickly than alternative approaches. Although ES has been shown to have superior performance on numerical problems \cite{hennig2012entropy},
investigating whether this claim holds true in practice
is future work.

A more challenging robotics scenario would be to use a consumer-grade vision system mounted on the head of the robot, instead of the current motion tracking system. This would provide observations of the pole state at a slower rate, larger measurement noise, and a potentially large delay. All these aspects may increase the cost function evaluation noise. It would be interesting to see how the system generalizes to these kinds of conditions, in future work.

\section*{ACKNOWLEDGMENT}
The authors thank Felix Grimminger and Heiko Ott for their support with the robot hardware and the pole design. We are grateful to Ludovic Righetti for his advice regarding the low-level tracking controllers, as well as Alexander Herzog for his help with the motion capture system.

\bibliographystyle{IEEEtran}
\bibliography{ICRA16}

\end{document}